\documentclass{article}

\usepackage{amsmath}
\usepackage{txfonts}
\usepackage{microtype}
\usepackage{graphicx}
\usepackage{subfig}
\usepackage{makecell}
\usepackage{booktabs} 

\usepackage{hyperref}

\usepackage[accepted]{icml2020}

\icmltitlerunning{A Meta Learning Approach to Discerning Causal Graph Structure}

\begin{document}

\twocolumn[
\icmltitle{A Meta Learning Approach to Discerning Causal Graph Structure}



\icmlsetsymbol{equal}{*}


\begin{icmlauthorlist}
\icmlauthor{Justin Wong}{equal,su}
\icmlauthor{Dominik Damjakob}{equal,su}
\end{icmlauthorlist}

\icmlaffiliation{su}{Department of Statistics, Stanford University, Palo Alto, United States}

\icmlcorrespondingauthor{Justin Wong}{juswong@stanford.edu}
\icmlcorrespondingauthor{Dominik Damjakob}{damjakob@stanford.edu}

\icmlkeywords{Machine Learning, Deep Bayesian Networks, Meta Learning}

\vskip 0.3in
]



\printAffiliationsAndNotice{\icmlEqualContribution} 

\begin{abstract}

We explore the usage of meta-learning to derive the causal direction between variables by optimizing over a measure of distribution simplicity. We incorporate a stochastic graph representation which includes latent variables and allows for more generalizability and graph structure expression. Our model is able to learn causal direction indicators for complex graph structures despite effects of latent confounders. Further, we explore robustness of our method with respect to violations of our distributional assumptions and data scarcity. Our model is particularly robust to modest data scarcity, but is less robust to distributional changes. By interpreting the model predictions as stochastic events, we propose a simple ensemble method classifier to reduce the outcome variability as an average of biased events. This methodology demonstrates ability to infer the existence as well as the direction of a causal relationship between data distributions.

\end{abstract}

\section{Introduction}
\subsection*{Background}
When it comes to inferring causal direction, the most popular tool is proper trial designs of experiments. More specifically, the randomized control trial (RCT) is a popular tool of choice since it allows us to easily separate treatment results from confounding variables so we are able to retrieve statistics such as average treatment effects that can inform the causality
 of a given treatment. 
However, such methods are not globally applicable since randomized trials can not be conducted in many scenarios: for example, they can be too costly, unethical or simply infeasible due to the complexity of real world systems.
Furthermore, the RCT is only applicable to prospective, but not retrospective studies - which is a large source of data to analyze. Applications in complex clinical trials or studies in social, economic and political sciences require more specialized tools to assist in discerning causality in the slew of data generated from less than ideal conditions in the modern computing era. 

Machine learning, most notably deep learning, is a powerful tool that has allowed for state-of-the-art performance in both discriminative  and generative tasks and has enjoyed huge amounts of growth in recent years as a result. However, canonical learning techniques often are likelihood based optimizations which converge regardless of causal direction with functionally equivalent parameterizations in both directions. 
Hence, we require specialized learning methods and importantly, additional assumptions that allow  deep learning models to discern the causal direction given their parameterizations.

We begin with the philosophy of Occam's razor - if multiple answers are correct, the best answer is the simplest one. Applied to our problem, this suggests that given the parameterizations $p(x | y)p(y)$ and $p(y | x)p(x)$ we assume that the "true" parameterization is the one that yields the simpler pair of conditional  and marginal distributions. Specifically, the measure of simplicity that we will use is the speed at which we are able to adapt a transfer distribution. Thus, under the assumptions, we expect that the correct causal direction will allow for faster transfer learning of the distribution under which we develop our methodology. 


\subsection*{Related Work}
While the derivation of causal directions with meta-learning is still a new research topic, first entries already exist. \citet{dasgupta2019causal} use meta-learning with model-free reinforcement learning to deduce causal reasoning in an end-to-end approach. \citet{bengio2019meta} instead apply meta-learning to derive the causal direction between variables in an optimization-based framework using Bayesian Networks. In their work, they apply learned models assuming different causal directions to data with a changed transfer distribution. As the correct causal model will only have to adjust its transfer distribution, and thus adapt faster, this allows it to extract the underlying causal directions. \citet{bengio2019meta} further apply this model to the Representation Learning domain, in which information from underlying variables has to be extracted. For this, they only consider a single freedom rotation framework as defined in \citet{bengio2013representation}, though more general models could provide better generalisation. 


The approach by \citet{CGNN} is a try at a more general approach through leverage a series of generative models in order to model each of the observable states of the graph. This allows it to resample a dataset distribution by sampling in topological order. Given a starting causal structure, their method refines the direction by an iterative method of resampling and computing a Maximum Mean Discrepancy (MMD) statistic that serves as a heuristic measurement as to the similarity of the ground truth distribution to the current iteration's resampled distribution. It then reassigns edge directions to reduce the MMD score and employs a cycle correcting method that allows it to resolve cyclic causal structure with its best acyclic counterpart by using a hill climbing technique. Here, the approach is flexible but rely on the correlation of MMD with correct causal structure in order to return a good assignment of the causal directions. 

\citet{metaCGNN} describe a meta-learning technique that is based on techniques from \citet{CGNN}. Noting computational issues of the original method, the authors propose a meta-learning method that re-frames assigning causal direction in a single graph as a meta-task. They then utilize a dataset of reference causal graphs and their corresponding meta-dataset to learn an efficient means of performing the meta-task. Thus, it is able to leverage similarity between meta-datasets to learn the causal direction of a new dataset more quickly. However, this method remains reliant on the usage of MMD and also follows the implicit assumption of knowing the existence of causal relationships and only needing to assign them direction. 

Several approaches exist to compute and train the log-likelihood of a model given a causal graph structure. The most popular are Variational Autoencoders \cite{varAutoEnc} and Bayes Networks \citet{DeepLearningBook}. Another approach is to parametrize the causal model more directly using Functional Causal Models (FCM) \cite{FCM}. FCMs further generalize results from \citet{CGNN} since we are no longer assuming that we know the exact distribution of latent confounders and instead allow stochasticity by way of inference through a proxy variable.



The contributions of this paper are that we are - by the authors best knowledge - the first to apply directional inference in meta-learning to variable structures with common confounders and improve on prior analysis by using FCMs. Additionally, we derive a more robust measure by introducing plurality voting and analyze our methodology for robustness properties to infer the feasibility of our approach.

\section{Methodology}
\subsection*{Meta Learning Causal Directions}
To learn the joint distribution of two variables $X$ and $Y$ we can use their conditional distributions $p_{x|y}$ and $p_{y|x}$ alongside their marginal distributions $p_x$ and $p_y$. 
In a Bayesian framework, this is expressed as 
\begin{flalign*}
    P_{X \rightarrow Y}(X, Y) = P_{X \rightarrow Y}(X) P_{X \rightarrow Y}(Y|X) \\
    P_{Y \rightarrow X}(X, Y) = P_{Y \rightarrow X}(Y) P_{Y \rightarrow X}(X|Y)
\end{flalign*}
where both parameterizations can be learned by Bayesian networks. Similar to \citet{bengio2019meta}, we assume that the true causal direction is $X \rightarrow Y$ and use the training distribution $p_0(x, y) = p_0(x) p(y|x)$. Thereafter, the distribution is changed to the transfer distribution $p_1(x, y) = p_1(x) p(y|x)$. Both networks are meta-trained to the transfer distribution for $T$ steps with resulting likelihoods 
\begin{flalign*}
    L_{X \rightarrow Y} = \prod_{t=1}^T P_{X \rightarrow Y, t}(x_t, y_t), \quad L_{Y \rightarrow X} = \prod_{t=1}^T P_{Y \rightarrow X, t}(x_t, y_t)
\end{flalign*}

which is trained in the following two step process:
\begin{enumerate}
    \item The relationship between $X$ and $Y$ is learned using two models: one assumes $X$ causes $Y$, the other the opposite causal direction.
    \item The distribution of $X$ is changed to a transfer distribution. Both models are retrained on the new data and the resulting likelihoods are recorded.
\end{enumerate}

Here, $P_{X \rightarrow Y, t}$ denotes the trained Bayesian network after step $t$.
Next, the loss statistic

\begin{equation*}
    R(\alpha) = -ln \left(\sigma(\alpha)L_{X \rightarrow Y}  + (1 - \sigma(\alpha)) L_{Y \rightarrow X} \right)
\end{equation*} 

is computed with $\alpha$ denoting a structural parameter defining the causal direction and $\sigma(\cdot)$ the sigmoid transformation. In this methodology, $\alpha$ is now optimized to minimize $R(\alpha)$. The loss statistic's gradient is 

\begin{equation*}
    \frac{\partial R}{\partial \alpha} = \sigma(\alpha) - \sigma(\alpha + ln(L_{X \rightarrow Y}) - ln(L_{Y \rightarrow X}))
\end{equation*}

such that $\frac{\partial R}{\partial \alpha} > 0$ if $L_{X \rightarrow Y} < L_{Y \rightarrow X}$, that is if $P_{X \rightarrow Y}$ is better at explaining the transfer distribution than $P_{Y \rightarrow X}$. \citet{bengio2019meta} show that if

\begin{equation*}
    E_{D_{transfer}}[ln(L_{X \rightarrow Y})] > E_{D_{transfer}}[ln(L_{Y \rightarrow X})]
\end{equation*}

where $D_{transfer}$ is the data drawn from the transfer distribution, stochastic gradient descent on $E_{D_{transfer}}[R]$ will converge to $\sigma(\alpha) = 1$ and $\sigma(\alpha) = 0$ if $ E_{D_{transfer}}[ln(L_{X \rightarrow Y})] < E_{D_{transfer}}[ln(L_{Y \rightarrow X})]$. As the loss function modeling the correct direction - this is, if $X$ causes $Y$, $L_{X \rightarrow Y}$ - only needs to update its estimate for the unconditional distribution $P_{X \rightarrow Y}(X)$ from $p_0(x)$ to $p_1(x)$ while the reverse direction networks needs to change both $P_{Y \rightarrow X}(Y)$ and $P_{Y \rightarrow X}(X|Y)$, it holds that indeed the loss statistic for the correct direction has a lower expected value and we can recover the causal direction.

\subsection*{Latent Variables Structure}

The previous results make the assumptions that the observed $X$ and $Y$ are independent of other hidden effects. However, this is unlikely to hold in practice. Noting the success of \citet{CGNN} in determining causal relations in graphs and their ability to express latent variables and effects directly, we adopt a similar representation of variable observations by using FCMs which suggest that observations are formed as tuples 
$$X_i = (\{P_i\}, f_i, \epsilon_i)$$
where $i$ indexes the vertex on the causal graph, $\{P_i\}$ denotes the set of causal parents of $X_i$, $\epsilon_i$ is independent noise modelling latent effects on $X_i$, and $f_i$ is a learned function.
\begin{figure}[ht]
\vskip 0.2in
\begin{center}
    \includegraphics[width=\columnwidth]{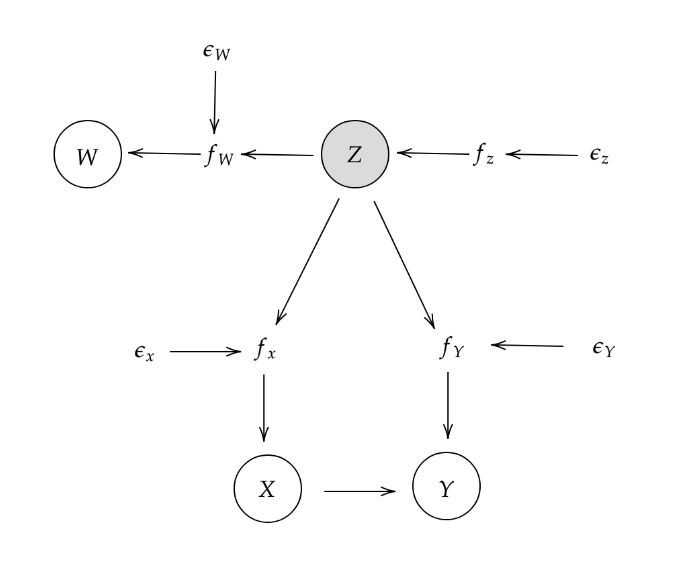}
    \caption{Example causal graph featuring observation variables $X$ and $Y$, latent confounder $Z$, and proxy variable $W$.}
    \label{fig: FCMPlot}
    \end{center}
    \vskip 0.0in
\end{figure}

As an example, in Figure \ref{fig: FCMPlot}, we can consider the FCM function for $Z$ to be $f_Z(\epsilon_Z)$ since it only has a causal parent in its unobserved, independent latent effects. In this case $\{P_Z\}$ is the empty set. To contrast, the FCM function for $Y$ would be $f_Y(X, Z, \epsilon_Y)$ since it has causal parents $X$, the latent $Z$, and its own independent hidden effects $\epsilon_Y$. In this case $\{P_Y\} = \{X, Z\}$. 

In the canonical definition of FCM, these functions look to predict the realization of the observable $\hat{X}_i = f_i(\{P_i\}, \epsilon_i)$. However, we find it more useful to use the FCM structure to predict model parameters for the distribution of the observable 
$$f_i: (\{P_i\}, \epsilon_i) \rightarrow (\{\pi_j\}_i, \{\mu_j\}_i, \{\sigma_j\}_i)$$
which we assume to be a Gaussian mixture with $j$ Gaussians. Then in our previously defined language, we have $p_{X | Y}(X | Y) = f_X(Y, \epsilon_X), P(Y) = f_Y(\epsilon_Y)$ which is the conditional and marginal distributions relevant for the $Y \rightarrow X$ direction and similarly  $P_{Y | X}(Y | X) = f_Y(X, \epsilon_Y), P(X) = f_X(\epsilon_X)$ are the relevant distributions for the $X \rightarrow Y$ direction. Given realizations of the ground truth observations, we then have a closed form for the likelihoods given each of the learned distributions.

\subsection*{Modelling Confounding Factors}
The basic FCM structure allows us to generalize the idea of an encoder-decoder structure by modelling hidden effects as different distributions $\epsilon_i$. Fortunately, it is also easy to extend to modelling latent confounders between variables. In particular, in the style of \citet{louizos2017causalLatent}, we introduce the latent variable $Z$ which can affect both $X$ and $Y$. Since $Z$ is not an independent hidden effect, it cannot be absorbed into either $\epsilon_X$ or $\epsilon_Y$. Instead, we must append $Z$ as an input to both $f_X$ and $f_Y$. 

While \citet{CGNN} assume that each confounder follows a known distribution, this is perhaps an overly ideal scenario that we are unlikely to encounter in practice. Instead, we choose to infer the values of $Z$. Let us consider the proxy variable $W$ that is also impacted by $Z$ such that $W|Z \perp \!\!\! \perp X|Z$ and $W|Z \perp \!\!\! \perp Y|Z$ as depicted in the setup from Figure \ref{fig: FCMPlot}. While $Z$ and $W$ can in principle be sets of variables of arbitrary length, for simplicity we restrict both to a single variable in this analysis. Further, we will model all causal causal effects of $Z$ on $X$, $Y$ and $W$ as additive effects.

To incorporate this variational inference of latent variables into the causal direction methodology, we follow \citet{variational} and assume that $Z$, $W|Z$ and $Y|X,Z$ are continuous, normally distributed variables with assumed
\begin{gather*}
    p(Z) \sim N(0, 1) \\
    p(W|Z) \sim N(\mu_W(Z), \sigma_W^2(Z)) \\
    p(X|Z) \sim N(\mu_X(Z), \sigma_X^2(Z)) \\
    p(Z|W) \sim N(\mu_Z(X), \sigma_Z^2(X))
\end{gather*}

We can estimate a lower bound for the combined probability of all variables by the Evidence Lower Bound (ELBO) which is defined by 
\begin{flalign*}
    L = \sum_{i=1}^N E_{p(z_i|x_i)}[&ln p(w_i|z_i) + ln p(x_i|z_i) + ln p(y_i|x_i, z_i) &\\
    & + ln p(z_i) - ln p(z_i|x_i)]
\end{flalign*}

We generate the ELBO for both causal directions such that they approximate $L_{X \rightarrow Y}$ and $L_{Y \rightarrow X}$. To compute the expected value inside the ELBO we use Monte Carlo simulation. Further, we use variational encoders to model $\mu_W(Z), \sigma_W^2(Z)$ and their alike parameters for $X$ and $Z$.


\subsection*{Super-runs and Plurality Voting}

Importantly, using ELBO instead of an exact likelihood introduces stochasticity since the bounds bounds can be of varying tightness for the different parameterizations. In particular, this holds for our approach as we use FCMs with specified error terms. As such, doing a single run is equivalent to observing a single outcome of a probabilistic event. When a causal direction exists, this probabilistic event is highly biased towards increasing/decreasing $\alpha$ depending on the direction, which can be observed in the robustness of results and consistency of $\alpha$ convergence. In contrast, when there is no causal direction, the probabilistic event is closer to random as demonstrated by highly varying $\alpha$ paths. Then while we gain an ability to perform inference of causal direction with latent variable structures, we are no longer able to accurately query existence of causal direction with a single result.

One simple and promising solution is to leverage multiple results in combination, called a super-run, to infer the existence and direction of causality. We consider perform a plurality vote to determine one of the three possible results. For each result, we consider the following parameters: 

\begin{enumerate}
    \item Voter count: The number of runs used
    \item Majority required: Percentage of the voter count required to win the election
    \item Cutoff: Values splitting the voting direction of an $\sigma(\alpha)$ path 
\end{enumerate}

Each run contributes a vote for either $X\rightarrow Y$ or $Y\rightarrow X$ when the $\sigma(\alpha\alpha)$ path cutoff threshold is made, otherwise the voter abstains. If one parameterization gains the designated majority required (i.e. 2/3 majority) then it is declared the winner. If neither achieves the required majority, then no causality is declared. 

It is also important to notice that by construction, performing an inference task is independent when conditioned on the data. Then the final $\sigma(\alpha)$ value falls within $[0, 1]$ with some distribution $f(\sigma(\alpha))$. Regardless of the distribution, we can summarize the probability of a positive vote as $p = \int_{\alpha_+}^1f(\alpha) d\alpha$ and similarly the probability of a negative vote as $q =  \int_0^{\alpha_-} f(\alpha) d\alpha$. Then we see that the super-run resulting in a decisive causal direction are tails of binomial random variables.
\begin{flalign*}
& \quad \quad \quad P(X\rightarrow Y) = \sum_{i = Nr}^N {N \choose i} p^{i}(1 - p)^{N - i} &\\
& \quad \quad \quad P(Y\rightarrow X) = \sum_{i = Nr}^N {N \choose i} q^{i}(1 - q)^{N - i} &\\
P(\textrm{$X\perp Y | Z$}) &= 1 -  \sum_{i = Nr}^N {N \choose i} p^{i}(1 - p)^{N - i} - \sum_{i = Nr}^N {N \choose i} q^{i}(1 - q)^{N - i}\\
&= 1 -  \sum_{i = Nr}^N {N \choose i} (pq)^{i}(1 - p - q + pq)^{N - i}
\end{flalign*}

These three quantities can further be simplified to a partition of a single binomial random variable if we take $\alpha_+ = \alpha_-$ to get $p = 1-q$.

\section{Experiments \& Results}
Unless otherwise specified, we modelled all data by the above process. Specifically, we modelled $Y$ as $Y \sim f(X) + Z$ where $f$ is a cubic spline function. For our experiments, we use 1000 observations for each draw of the training and transfer distribution, as this is also the number used by \cite{bengio2019meta}, and 300 iterations of the Monte Carlo method. The trained models are meta-learned for 5 steps and the FCM uses 5 Gaussians. 

Moreover, we model $X$ by $X \sim N(\mu_X, 2) + Z$
with $\mu_X = 0$ for the training distribution and $\mu_X \sim U(-4, 4)$ under the transfer distribution. For our inference analysis we assumed that all variables are normally distributed. All causal models are trained for 500 iterations and the $alpha$ parameter for 400 iterations and the results for $\sigma(\hat{\alpha})$ are extracted in the end. 

\subsection*{Normality Results}
\begin{figure}[ht]
\vskip 0.2in
\begin{center}
    \centering
    \subfloat[a][Results for $\sigma(\alpha)$]{
       \includegraphics[width=\columnwidth]{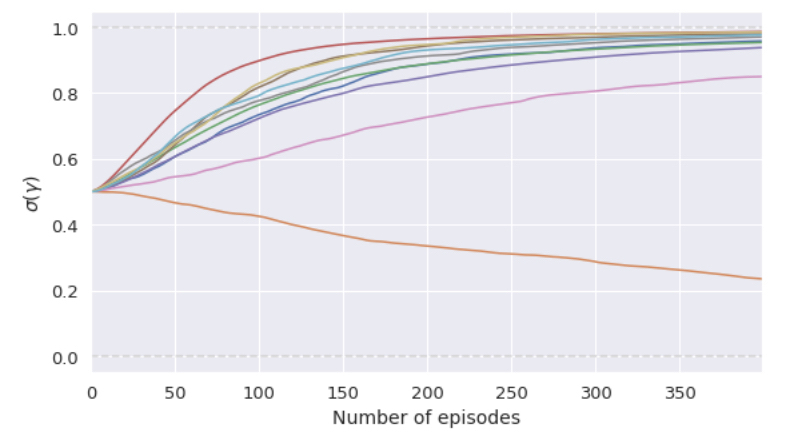}
       \label{fig: Normal_alpha}}
       
    \centering
    \subfloat[a][Results for $\sigma(\alpha)$]{
       \includegraphics[width=\columnwidth]{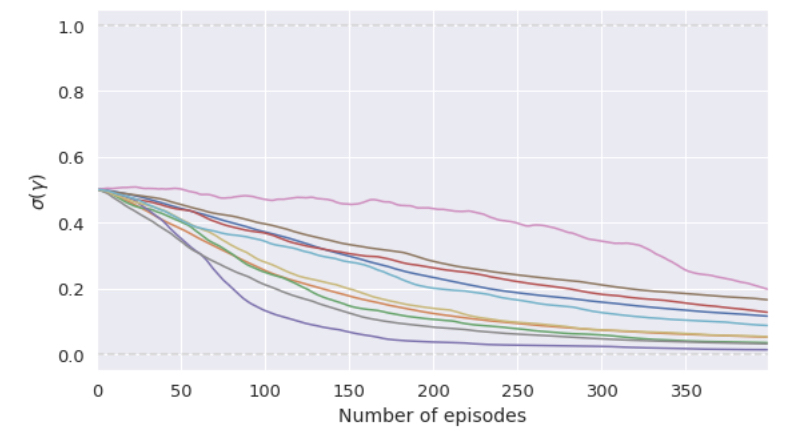}
       \label{fig: reverse_alpha}}
    \caption{$\sigma(\alpha)$ estimates for the the standard model. We generate two models with opposite directions. In \ref{fig: Normal_alpha} $X$ causes $Y$ while in \ref{fig: reverse_alpha} $Y$ is the true cause of $X$. Both plots show 10 different optimization curves for distinct i.i.d. data}
    \label{fig: NormalPlots}
    \end{center}
    \vskip 0.0in
\end{figure}

To show the functionality of our methodology a plot of the estimation path for $\sigma(\alpha)$ with 10 repetitions is provided in Figure \ref{fig: BetaPlots}. The graphs depicts two distinct scenarios: in the upper plot our true model states that $X$ causes $Y$ while in Figure \ref{fig: reverse_alpha} we model that $Y$ causes $X$. This allows us to evaluate the properties of our optimisation method without falling for a potential fallacy if $\alpha$ is biased in a certain direction. 

The sigmoid transformation of $\alpha$ shows that in both cases we are able to infer the correct causal direction. For the model that specifies $X$ causing $Y$ we can observe that in 9 out of 10 cases $\sigma(\alpha)$ increases beyond 0.8 and in fact converges towards 1 for larger epochs in many such cases, while for the reverse causal direction $\sigma(\alpha)$ decreases towards 0 in all 10 instances.

We can further observe that for most instances $\sigma(\alpha)$ goes towards 0 relatively fast and the function has already converged after about 250 iterations. Therefore, this analysis supports the overall feasibility of our methodology.

\subsection*{Analysis of the FCM}
To analyze the correctness of our results we can investigate the output of the FCM, specifically the estimate for the mean of the predicted variable. If our assumed causal direction is that $X$ causes $Y$ and the FCM models $k$ Gaussian variables then we can use  
\begin{equation*}
    \hat{y}(x, z) = \frac{\sum_{i=1}^k \pi_{i | x,z} \mu_{i | x,z}}{\sum_{i=1}^k \pi_{i | x,z}}
\end{equation*}
as the prediction of the mean of $Y$ given $X$ and $Z$. The FCM mean of $X$ can be inferred in a likewise fashion. For this analysis, we use $X=0$, $Y=0$ and $Z = 0$ as input variables to predict the conditioned mean. 

\begin{figure}[ht]
\vskip 0.2in
\begin{center}
    \centering
    \subfloat[a][Results for the model assuming $X$ causes $Y$]{
       \includegraphics[width=\columnwidth]{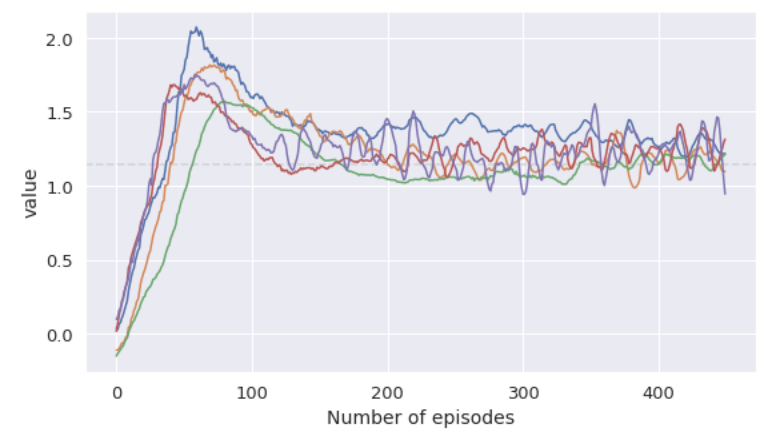}
       \label{fig: FCM_ouput_standard_x2y}}
       
    \centering
    \subfloat[a][Results for the model assuming $Y$ causes $X$]{
       \includegraphics[width=\columnwidth]{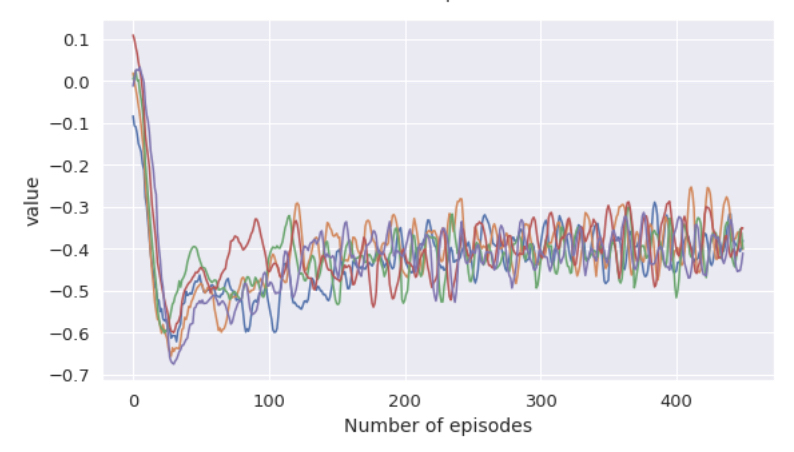}
       \label{fig: FCM_ouput_standard_y2x}}
    \caption{output of the FCM indicating the mean of $Y$ given $X$ and $Z$ for $X = 0, Z = 0$. The dashed line shows the actual value of $Y$ at this point}
    \label{fig: FCM_ouput_standard}
    \end{center}
    \vskip 0.0in
\end{figure}

Plots for this experiment for 10 repetitions can be seen in Figure \ref{fig: FCM_ouput_standard}. For the causal direction $X \rightarrow Y$ in Figure \ref{fig: FCM_ouput_standard_x2y} the dashed line indicates the actual mean of $Y|X=0, Z=0$ for our spline function. As the spline does not have a proper inverse function, no such line is included in Figure \ref{fig: FCM_ouput_standard_y2x}.

As Figure \ref{fig: FCM_ouput_standard_x2y} shows, the FCM mean converges to the correct conditioned mean for all ten repetitions. For all iterations this happens within the first 200 episodes. As we only use our converged FCM when training $\alpha$ and $\beta$, this shows that all transfer distributions will use a correct FCM in the computation of their ELBOs. Additionally, we can also note that in Figure \ref{fig: FCM_ouput_standard_y2x} the conditional means also move around the same conditional mean, such that their models also appear to have converged.

In models that train correctly, we expect that the total variation after convergence of the FCM should be due to the $\epsilon$ noise. Seeing that the FCM models for $X\rightarrow Y$ converge to the same variance after many iterations adds evidence towards that causal direction. In contrast, the variance of $Y\rightarrow X$ FCM models fail to consistently attain the same variance near the end of training which suggests that indeed this is the wrong causal direction. 

The reason for the more erratic behaviour for the Y causes X direction is s the aforementioned non-invertibility of the spline function. As several X-values correspond to $y = 0$, the models fluctuate around the conditional mean of these values.



\subsection*{Detection of no Causality}

\begin{figure}[ht]
\vskip 0.2in
\begin{center}
\centerline{\includegraphics[width=\columnwidth]{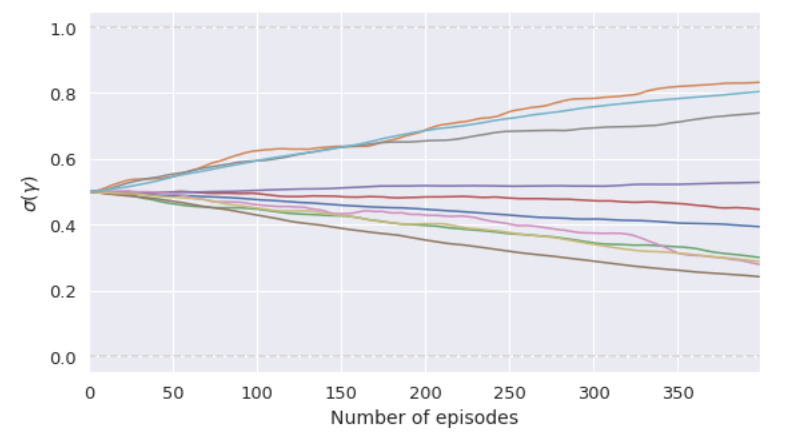}}
\caption{10 repetitions of $\sigma(\alpha)$ paths for scenario with no causal relationship}
\label{fig: UncorrPlots}
\end{center}
\vskip -0.2in
\end{figure}

To demonstrate the behavior of our model for the case where there is no causal relationship between $X$ and $Y$ variables, we consider a independent ground truth relationship after conditioning on the latent variable $Z$. In Figure \ref{fig: UncorrPlots} we plot the paths for $\sigma(\alpha)$ in this problem setup, again for 10 repetitions.

The different directions of the sigmoid transformation for $\alpha$ indicate that the model does not find a clear causal direction in this scenario. While some of the curves increase or decrease towards $1$ or $0$, $\sigma(\alpha)$ remains at 0.5 for many instances. Therefore, our model is able to infer for such scenarios that no causal direction yields a lower loss function and thus indicates that there is no causal relationship. Further, for cases in which the parameter does not remain flat, the optimization curves do not show drastic changes but instead increase or decrease only slowly when compared to the prior analysis in Figure \ref{fig: NormalPlots} that contained an existing causal structure. Hence, this analysis supports the robustness of our analysis.


\subsection*{Super-runs}
For these experiments, we use two symmetric cutoffs. The first is $\alpha_+ = \alpha_- = 0.5$ and the second is $\alpha_+ = 0.7 \neq \alpha_- = 0.3$, both with $N = 10$. We perform two experiments with each configuration, one where there is causality $X\rightarrow Y$ and one where there is no causal relation. 

With $\alpha_+ = \alpha_- = 0.5$ and under $X\rightarrow Y$, we find that the super-run concludes a the correct causal relationship all ten times. Under no causal relationship, the super-run concludes correctly that there is no causal relationship seven times and predicts $X\rightarrow Y$ three times.

With $\alpha_+ = 0.7, \alpha_- = 0.3$ and under $X\rightarrow Y$, we find that the super-run concludes a the correct causal relationship eight times and concludes no causal relationship twice. Under no causal relationship, the super-run concludes correctly that there is no causal relationship ten times.


Clearly, choosing the cutoff point is a hyperparameter tuning problem that can be optimized to find the desired sensitivity to errors for both causal discovery and causal direction. It is also easy to improve results by adding more voters. These additional voters can be run in parallel such that a super-run can be achieved in the same time as a single result given sufficient cores. We further note that the existence of causality can be established using standard causal inference methods such that a super-run with $\alpha_+ = \alpha_- = 0.5$ can be applied to identified relationships to only detect the direction of the causality.

\subsection*{Robustness for Non-normality}
\begin{figure}[ht]
\vskip 0.2in
\begin{center}
    \centering
    \subfloat[a][Results for $Z \sim Beta(0.5, 0.5)$]{
       \includegraphics[width=\columnwidth]{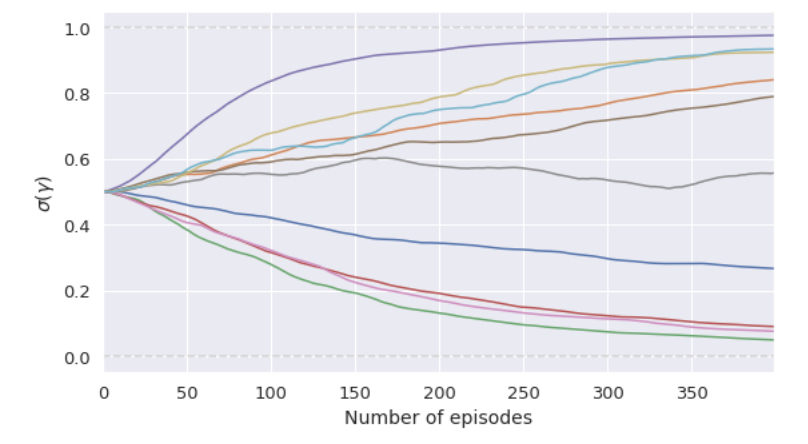}
       \label{fig: Beta_alpha}}
       
    \centering
    \subfloat[a][Results for $Z \sim Beta(3,3)$]{
       \includegraphics[width=\columnwidth]{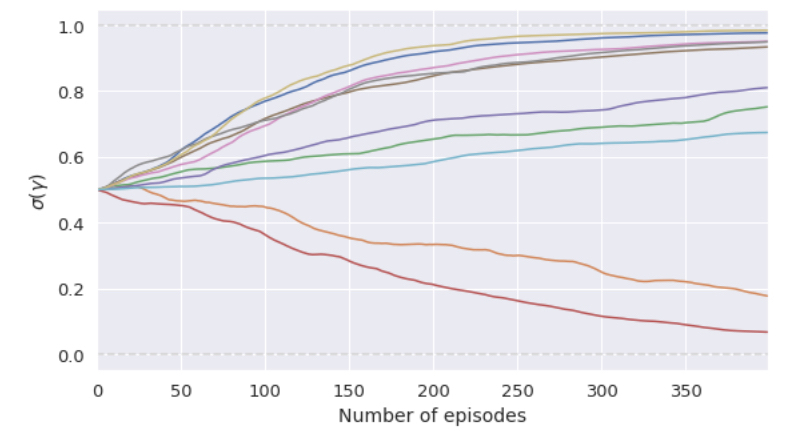}
       \label{fig: Beta_beta}}
    \caption{$\sigma(\alpha)$ estimates if $Z \sim Beta(0.5, 0.5)$ or $Beta(3,3)$.}
    \label{fig: BetaPlots}
    \end{center}
    \vskip 0.0in
\end{figure}

In our model we correctly assume that $Z$ is normally distributed when estimating its parameter for our causal inference. As this assumption of a correct causal model is not always applicable in practice it is of interest if our inference results keep their predictive power if this is not the case. Hence, we model $Z$ as a Beta distributed variable centered around 0 with parameters $a$ and $b$ and choose 2 variable combinations for the two variables. The first is $0.5, 0.5$, which has a U-shaped distribution function and is therefore strongly dissimilar to a Normal distribution. The second is $a=3, b=3$, which has a more similar probability density function to the Normal distribution. The results for the two distributions are shown in Figure \ref{fig: BetaPlots}.
As can be seen, for both distributions the methodology performs worse in predicting the correct causal relationship. For the $Beta(0.5, 0.5)$ distribution, the model can no longer reliably predict the correct causal direction and only depicts correct convergence behavior in 5 out of 10 instances. For the $Beta(3,3)$ distribution however, we retain our ability to deduce the correct causal direction, with 8 correct out of 10 predictions. Yet, we can also observe that our model takes longer to converge to the correct value of $\sigma(\alpha) = 1$ even in this scenario. This is an additional indicator that the accuracy of the model was reduced under the new distribution. Overall, we hypothesize that this decrease in predictive performance occurs as the used ELBO is not a strong enough bound and too dependent on the assumed distribution of the latent variable. This issue could be overcome by constructing a more robust ELBO function, which we leave to further research. 


\subsection*{Robustness for Limited Sample Data}

In previous results, we sample data from well defined ground truth distributions. In particular, for each iteration step of the training process we sample new data from the specified distributions and - in case of the training distribution - randomly change the mean of $X$'s probability distribution to derive the transfer distribution. However, in real world applications, there will only exist a finite dataset - some number of realizations from the ground truth distribution which is inaccessible and only a limited number of transfer distributions. In effect, by sampling from the ground truth directly, we are assuming some infinite size dataset from which we draw data. Instead, a more realistic approach is to mimic a finite dataset environment by first creating a static dataset. Afterwards, we can draw data samples from it during training and compute our parameter estimates in the usual way. For such a more realistic approach it is also important to account for the effect that in practice there exists only a limited number of transfer distributions.

Therefore, to define these finite datasets we use four hyperparameters:
\begin{enumerate}
    \item The total number of observations in the training distribution 
    \item The total number of observations in each sample of the transfer distribution
    \item The number of observations used in each training episode
    \item The number of distinct transfer distributions
\end{enumerate}
When sampling our data from the predefined datasets we only use a portion of that data during each training episode. Specifically, we will use the Bootstrap to sample our data, such that we draw samples with replacement from the datasets.

Table \ref{Tab:LimitTable} depicts mean and standard deviation results for $\sigma(\alpha)$ over 10 iterations each for different combinations of the four hyperparameters. To analyze effects for datasets of different size we used 1000, 100 and 30 total observations for the training samples and 500, 50 and 15 for each of the transfer distributions. For each combination, we used a fixed amount of observations per training episode and 10, 3, and 1 transfer distributions. 

\begin{table*}[ht]
\centering  
\begin{tabular}{c|c|c|c|c|c} 
\thead{Total observations \\ for training sample} & \thead{Total observations \\ for transfer samples} & \thead{Observations \\ per episode} & \thead{Total transfer \\ distributions} & Mean of $\sigma(\alpha)$ & St.d. of  $\sigma(\alpha)$\\
\hline
1000               & 500                                     & 200                      & 10                               & 0.995      & 0.009    \\
1000               & 500                                     & 200                      & 3                               & 0.823      & 0.306   \\
1000               & 500                                     & 200                     & 1                                & 0.849      & 0.235    \\
100                & 50                                     & 50                       & 10                               & 0.898      & 0.294    \\
100                & 50                                     & 50                       & 3                                & 0.801      & 0.395    \\
100                & 50                                     & 50                       & 1                                & 0.800      & 0.396    \\
30                & 13                                     & 13                       & 10                                & 0.300      & 0.458    \\
30                & 15                                     & 15                       & 3                                & 0.500      & 0.500    \\
30                & 15                                     & 15                       & 1                                & 0.500      & 0.500
\end{tabular}
\caption{Mean results for predefined datasets. For all results we used the average of 10 repetitions.}
\label{Tab:LimitTable}
\end{table*}

The table shows that for strongly restricted preselected data samples the predictive accuracy of our methodology decreases, while no strong effect can be detected for moderate restrictions. For example, for 1000 training samples with 10 transfer distributions we reach a mean value of 0.995 for $\sigma(\alpha)$ which drops to 0.300 for 30 samples. Therefore, our methodology works with moderately limited data sets and can therefore be used in practical applications

Moreover, the effect of stark reductions of the number of available training sets are also distorting. While we remain able to identify the correct causal direction for 10 transfer distributions, this capability is greatly limited for lower values. For instance, for 1000 training and 500 transfer observations with 3 transfer distributions the mean of $\sigma(\alpha)$ drops to 0.823. Our explanation is that for a limited number of transfer sets the optimized parameter for $\alpha$ will strongly depend on the randomly chosen value for $\mu$, which determines the mean of the transfer distribution. If a value close to 0 is selected, the training and transfer sets are not sufficiently different to guarantee a appropriate difference of the ELBOs. Therefore, under the assumption that the data generating processes guarantee a sufficient difference of the data sets, this effect may be less important in practice. 



\section{Conclusions}

In our experiments we have improved on the the meta-learning approach from \citet{bengio2019meta} and \citet{dasgupta2019causal} to express the causal graph structure more explicitly by using FCM. This innovation yields great improvements in the results as we are able to demonstrate faster and better convergence to higher confidence of the correct causal direction in more difficult problem setups as compared to prior works. In particular, we have shown that for generalized independent hidden effects and with single latent confounders, we are able to recover the correct causal direction with high confidence. The framework that we use is easily extendable to larger number of confounding factors as well as more observational variables as they can be fit into our structure by introducing more inputs to the relevant FCM networks and more FCM networks to learn respectively. 

Our analysis further shows that our architecture is also able to predict the existence of a causal relationship. Especially with the introduction of super-runs that combine several model predictions using the notion of conditional independence both our predictive performance of the correct causal direction and its existence improve, though some caution with respect to the cutoff parameter should be kept.


Finally, we show that in cases where model assumptions are violated, the model's predictive performance decreases, but the model is robust overall with respect to moderate violations. When we have distributional deviation from normality, the $\alpha$ paths have more difficulty converging but still tend towards the correct direction if the actual distribution resembles a normal distribution. When we constrain the model to finite datasets, the probability of converging to the correct causal relation decreases as a function of dataset size and the number of transfer distributions, but we empirically still see that we can often recover the correct relation regardless. 

\subsection*{Directions for Further Research}

In order to generalize assumptions about the distribution of confounding effects, we introduce a variational inference framework for the latent variables that we measure through some proxy variable. However, this also necessitates the replacement of the exact likelihood with the ELBO
which is not a sharp bound and allows that $L_{X\rightarrow Y} > L_{Y \rightarrow X}$ while also having $ELBO_{X \rightarrow Y} < ELBO_{Y \rightarrow X}$. Especially for the non-normality tests, this constituted a major issue. 
Therefore, we advise that additional work on this implementation of the ELBO could find a more robust bound and thus improves convergence towards the correct causal direction.


Furthermore, this work can be extended to larger causal graphs with more complex latent and observed variable structures. To infer the causal direction on larger graphs, we advise to assume a starting orientation and iteratively perform a two step process similar to the method of \citet{CGNN} to perform hill climbing.

\begin{enumerate}
    \item Update causal direction scheme by relearning the transfer distributions in topological order using the current orientation to decide the causal parents of variables.
    \item Resolve any formed cycles by reorienting violating causal arrows with the smallest confidence
\end{enumerate}

However, we notice the ability of meta-CGNN in \citet{metaCGNN} to use dataset and causal direction pairs as meta-tasks to train networks that can leverage dataset similarities to find causal directions during meta-test time more quickly. It may be possible to combine these two works in order to consolidate the computational complexity of extending work to larger graph sizes.

Finally, we explore only some perturbations to the distributional assumptions. However, there are a larger number of possibilities to explore as well as extensions to more complex datasets. Improvements on these fronts will allow for the development of agents with improved robustness and usefulness in practice.

\appendix

\bibliography{biblio}
\bibliographystyle{icml2020}

\end{document}